\pdfoutput=1

\documentclass[11pt]{article}

\usepackage[]{EMNLP2022}

\usepackage{times}
\usepackage{latexsym}

\usepackage{booktabs,subcaption,amsfonts,dcolumn}
\usepackage{adjustbox}

\newcommand\tf[1]{\textbf{#1}}

\usepackage[T1]{fontenc}

\usepackage[utf8]{inputenc}

\usepackage{microtype}

\usepackage{inconsolata}

\usepackage{array}
\usepackage{pifont}
\usepackage{tabularx}
\usepackage{adjustbox}
\usepackage{multirow}
\usepackage{enumitem}
\usepackage{xspace}
\usepackage{tcolorbox}
\usepackage{booktabs,subcaption,amsfonts,dcolumn}
\usepackage{url}
\usepackage{amsmath,amsthm,amsfonts,amssymb,bm,stmaryrd}
\usepackage{color}
\usepackage{CJKutf8}
\usepackage{makecell}
\usepackage{booktabs}
\usepackage{graphicx}

\usepackage{array}
\usepackage{pifont}
\usepackage{tabularx}
\usepackage{adjustbox}
\usepackage{multirow}
\usepackage{enumitem}
\usepackage{xspace}
\usepackage{tcolorbox}
\usepackage{booktabs,subcaption,amsfonts,dcolumn}
\usepackage{url}
\usepackage{amsmath,amsthm,amsfonts,amssymb,bm,stmaryrd}

\title{Contrastive Demonstration Tuning for Pre-trained Language Models}

\author{
Xiaozhuan Liang$^{1,2}$, 
Ningyu Zhang$^{1,2}\thanks{\quad Corresponding author}$, 
Siyuan Cheng$^{1,2}$,
Zhenru Zhang$^{3}$, \\
\textbf{Chuanqi Tan$^{3}$, 
Huajun Chen$^{1,2}$}  \\
	$^1$Zhejiang University \& AZFT Joint Lab for Knowledge Engine, China \\
	$^2$Hangzhou Innovation Center, Zhejiang University, China \\
	$^3$Alibaba Group, China \\
 {\tt \{liangxiaozhuan,zhangningyu,sycheng,huajunsir\}@zju.edu.cn}\\
 {\tt \{zhangzhenru.zzr,chuanqi.tcq\}@alibaba-inc.com} 
}

\begin{document}
\maketitle
\begin{abstract}
 
Pretrained language models can be effectively stimulated by textual prompts or demonstrations, especially in low-data scenarios. Recent works have focused on automatically searching discrete or continuous prompts or optimized verbalizers, yet studies for the demonstration are still limited. Concretely, the demonstration examples are crucial for an excellent final performance of prompt-tuning. In this paper, we propose a novel pluggable, extensible, and efficient approach named contrastive demonstration tuning, which is free of demonstration sampling. Furthermore, the proposed approach can be: (i) Plugged into any previous prompt-tuning approaches; (ii) Extended to widespread classification tasks with a large number of categories. Experimental results on 16 datasets illustrate that our method integrated with previous approaches LM-BFF and P-tuning can yield better performance\footnote{Code is available in \url{https://github.com/zjunlp/PromptKG/tree/main/research/Demo-Tuning}.}.

\end{abstract}

\section{Introduction}


%


Pre-trained language models (PLMs) have been applied to widespread natural language understanding and generation tasks, which are proven to obtain significant gains across benchmarks \cite{DBLP:conf/naacl/DevlinCLT19,DBLP:journals/corr/abs-1907-11692,DBLP:conf/acl/LewisLGGMLSZ20,DBLP:conf/nips/00040WWLWGZH19,DBLP:conf/icml/Bao0WW0L0GP0H20,DBLP:journals/corr/abs-2201-05575,DBLP:journals/corr/abs-2202-02113,DBLP:journals/corr/abs-2201-11147}. 
One paradigm of PLMs is the pre-train—fine-tune, which has become the \emph{de facto} standard for natural language processing (NLP), where task-specific objectives and additional parameters are leveraged in the tuning procedure. 
Recently, the paradigm of the adaptation of PLMs has been shifting.
A new fine-tuning methodology named prompt-tuning with a natural language \textbf{prompt} and a few \textbf{demonstrations} has made waves in the NLP community by proving astounding few-shot capabilities on myriad language understanding tasks.
Further studies try to mitigate the labour-intensive prompt engineering with discrete prompt searching \cite{DBLP:conf/emnlp/ShinRLWS20} or continuous prompt optimization   \cite{DBLP:journals/corr/abs-2103-10385,DBLP:conf/acl/LiL20,DBLP:conf/acl/HambardzumyanKM20,DBLP:conf/naacl/ZhongFC21}. 
However, few studies have focused on the demonstration, which is an indispensable component in prompt-oriented methodologies.

 \begin{figure}[!t]
     \centering
     \includegraphics[width=1.0\columnwidth]{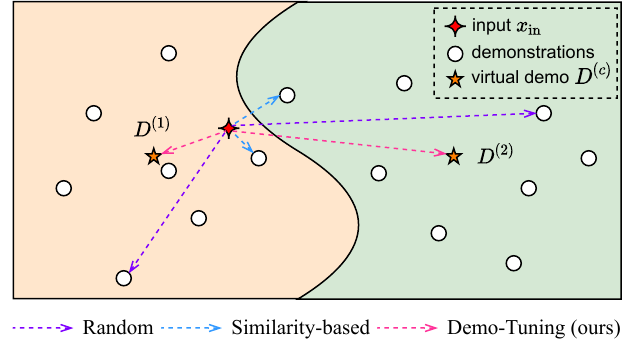}
     \caption{Comparison among current sampling strategies on demonstration-based learning. Compared to random and similarity-based sampling, demo-tuning can obtain better demonstration distributions.}
     \label{fig:motivation}
 \end{figure}

In previous studies, demonstrations are sampled examples in the training set.
GPT-3’s naive “in-context learning” paradigm picks up to 32 randomly sampled instances as demonstrations and directly concatenates them with the input sequence \cite{DBLP:journals/corr/abs-2101-06804,rethink-demo}.
Since informative demonstrations are crucial for model performance, \citet{DBLP:conf/acl/GaoFC20} develop a refined strategy via sampling input pairs with similar examples, thereby providing the model with more discriminative comparisons.
However, it is still not guaranteed to prioritize the most informative demonstrations as (1) the similarity-based sampling may obtain degraded demonstrations in different classes but have similar distances to the input; (2) the number of usable demonstrations is still bounded by the model’s maximum input length.
For example, as shown in Figure \ref{fig:motivation}, the purple lines refer to the random sampling while the blue lines indicate similarity-based sampling. 
Note that similarity-based sampling may obtain examples very similar to the input sequence.
However, those sampled examples with different labels may tend to have a similar representation and thus confuse the discriminability of the model.
Moreover, for datasets with many classes, it is still non-trivial to concatenate all sampled demonstrations. 
Those above-mentioned challenges hinder the applicability of demonstration in prompt-tuning. 

To address those issues, in this paper, we propose contrastive \textbf{DEMO}nstration \textbf{Tuning} (Demo-tuning) for pre-trained language models. 
Specifically, we leverage learnable continuous embeddings (e.g., one or two learnable tokens) as virtual demonstrations to relax the maximum number of categories. 
We concatenate those virtual demonstrations to the input sequence; thus, our approach can be extended to a wide variety of classification tasks with many categories.
To optimize those continuous embeddings, we explore a simple contrastive framework without negative pairs \cite{DBLP:conf/nips/GrillSATRBDPGAP20} since it is difficult to find an appropriate negative pair in semantic space for NLP.
In each training batch, we randomly sample a real example and regard the virtual and real examples as positive pairs. 
With contrastive learning, we can obtain informative, optimized virtual demonstrations with more discriminative comparisons.

We conduct extensive experiments on 16 NLP datasets.
Our contrastive demonstration tuning can yield better performance when integrated with previous prompt-based methods (e.g., LM-BFF \cite{DBLP:conf/acl/GaoFC20}, P-tuning \cite{DBLP:journals/corr/abs-2103-10385}).
Moreover, our approach can be applied to datasets with many categories and outperform baselines.
Note that our approach is model-agnostic and can be plugged into lots of prompt-based methods without the effort to select suitable demonstrations.
The main contributions of this study are as follows:

\begin{itemize}
    \item We propose a pluggable, extensible, and efficient approach to contrastive demonstration tuning for pre-trained language models. 
    To the best of our knowledge, optimizing demonstration is also a new branch of research that has not been explored in language model prompting.
    
    \item We propose virtual demonstration and leverage contrastive learning to obtain informative demonstrations and also relax the maximum number of categories in classification tasks.
    
    \item A systematic evaluation of 16 NLP datasets shows that the proposed simple-yet-effective approach contributes towards improvements across all these tasks.  
\end{itemize}

\section{Related Work}

\subsection{Prompt-tuning}

With the prevalence of GPT-3 \cite{DBLP:conf/nips/BrownMRSKDNSSAA20}, prompting PLMs for few-shot learning has become a new, popular learning paradigm   in natural language processing \cite{DBLP:conf/eacl/SchickS21,DBLP:journals/corr/abs-2103-11955,DBLP:journals/corr/abs-2107-13586} and appealed to researchers.
Recently, prompt-tuning has been applied to various NLP tasks, such as named entity recognition \cite{DBLP:conf/acl/CuiWLYZ21,chen2021lightner,DBLP:journals/corr/abs-2110-07331, DBLP:conf/naacl/MaZGTLZH22}, entity typing \cite{DBLP:journals/corr/abs-2108-10604}, relation extraction \cite{DBLP:journals/corr/abs-2105-11259}, event extraction \cite{hsu2021event,ye2021learning}, sentiment analysis \cite{li2021sentiprompt}, machine translation \cite{DBLP:journals/corr/abs-2110-06609}, and knowledge graph completion \cite{xie2022discrimination}. 
\citet{DBLP:conf/eacl/SchickS21,DBLP:journals/corr/abs-2009-07118} propose the PET, which reformulates the NLP tasks as cloze-style questions and yields satisfactory performance.
\citet{DBLP:journals/corr/abs-2103-11955} further propose a denser supervision object during fine-tuning to improve the PET. 

Note that handcrafting a best-performing prompt is like finding a needle in a haystack,  which facilitates the labor-intensive prompt engineering, 
Thus, recent studies \cite{DBLP:journals/corr/abs-2104-06599,DBLP:journals/corr/abs-2101-00121,DBLP:journals/corr/abs-2201-11332,DBLP:journals/corr/abs-2104-07650} conducted in this field have been focused on automatically searching the prompts. 
\citet{DBLP:conf/emnlp/ShinRLWS20}  propose AUTOPROMPT, which is a gradient-based method to acquire templates and label words for prompt-tuning. 
\citet{DBLP:journals/corr/abs-2104-14690} propose  EFL, which reformulates the NLP task as an entailment one and turns small LMs into better few-shot learners. 
Additionally, \citet{DBLP:journals/corr/abs-2012-15723} propose LM-BFF—better few-shot fine-tuning of language models, which utilizes a generation model to obtain templates and a refined strategy for dynamically and selectively incorporating demonstrations into each context. However, it is sub-optimal for the discrete prompt searching due to the continuous nature of neural networks. 

To overcome these limitations, \citet{DBLP:journals/corr/abs-2103-10385,DBLP:journals/corr/abs-2110-07602} propose P-tuning to to automatically search prompts in the continuous space. 
\citet{DBLP:conf/acl/LiL20} propose prefix-tuning, which optimizes a sequence of continuous task-specific vectors and keeps language model parameters frozen.
\citet{DBLP:journals/corr/abs-2104-08691} leverage a mechanism to learn “soft prompts” to condition frozen language models.
\citet{DBLP:journals/corr/abs-2108-13161} propose a differentiable prompt learning method for few-shot NLP with optimized prompt templates as well as labels. 
\citet{DBLP:journals/corr/abs-2110-07904} propose SPoT, which learns a prompt on one or more source tasks and then uses it to initialize the prompt for a target task to boost the performance across many tasks.
More related works including WARP \cite{DBLP:conf/acl/HambardzumyanKM20} and OPTIPROMPT \cite{DBLP:conf/naacl/ZhongFC21} also propose to  leverage continuous templates, which is more effective than discrete prompt search.
To conclude, most of the existing works try to obtain optimized prompts for widespread NLP tasks; however, few studies have focused on the demonstration, which is an indispensable component in prompt-oriented learning. 

Our work is orthogonal to previous prompt-tuning approaches, which are aimed at optimizing prompts. 
The major differences between virtual demonstration and continuous prompts are that:
1) they have a wholly different training strategy since continuous prompts are optimized via backpropagation with a training set, while our approach utilizes contrastive learning. 
2) our approach requires no external architecture (e.g., LSTM in P-tuning), thus, making it efficient and pluggable to any prompt-tuning approaches.
To date, \citet{DBLP:journals/corr/abs-2110-08454} is the only approach that studies the demonstration and presents a simple demonstration-based learning method for named entity recognition. 
Apart from \citet{DBLP:journals/corr/abs-2110-08454}, our approach focus on general NLP classification tasks. 
Moreover, we propose virtual demonstrations with contrastive learning strategies, which can obtain better demonstrations and also relax the maximum number of categories in datasets. 

\subsection{Contrastive Learning}

Contrastive learning has been long considered effective in learning meaningful representations. 
In the early stage, \citet{DBLP:conf/nips/MikolovSCCD13} propose to learn word embeddings by regarding words nearby a target word as a positive instance while others as negative. 
\citet{DBLP:conf/iclr/LogeswaranL18,DBLP:conf/cicai/ChenXBYDZC21} further generalize this approach to learn  sentence representations. 
Recently, \citet{DBLP:conf/acl/KimYL20}  propose a contrastive learning method that makes use of a  self-guidance mechanism.
\citet{DBLP:conf/acl/YanLWZWX20} propose ConSERT, a contrastive framework for self-supervised sentence representation transfer.
\citet{DBLP:conf/acl/GiorgiNWB20}  propose  DeCLUTR: Deep Contrastive
Learning for Unsupervised Textual Representations. 
\citet{DBLP:conf/emnlp/GaoYC21} leverage dropout as mimimal data augmentation and propose SimCSE, a simple contrastive learning framework that greatly advances the state-of-the-art sentence embeddings.

On the other hand, contrastive learning has been also appealed to the computer vision community \cite{DBLP:journals/corr/abs-2011-00362,DBLP:journals/corr/abs-2006-08218}.  
\citet{DBLP:conf/icml/ChenK0H20} propose SimCLR: a simple framework
for contrastive learning of visual representations without requiring
specialized architectures or a memory bank. 
\citet{DBLP:conf/cvpr/ChenH21} observe that simple siamese networks can learn meaningful representations even using none of the negative sample pairs, large batches, and momentum encoders.
 
Our work is related to \citet{DBLP:conf/nips/GrillSATRBDPGAP20}, a non-contrastive self-supervised learning approach, which relies on two neural networks, referred to as online and target networks, that interact and learn from each other.
However, as opposed to this approach, we utilize the encoder in the same state while \citet{DBLP:conf/nips/GrillSATRBDPGAP20} leverage two networks in the different states.
Moreover, we focus on demonstration optimization in prompt-tuning for NLP, including learning informative demonstrations and acquiring prompt temples and label tokens. 
 
 \begin{figure*}[t]
    \centering
    \includegraphics[width=0.95\textwidth]{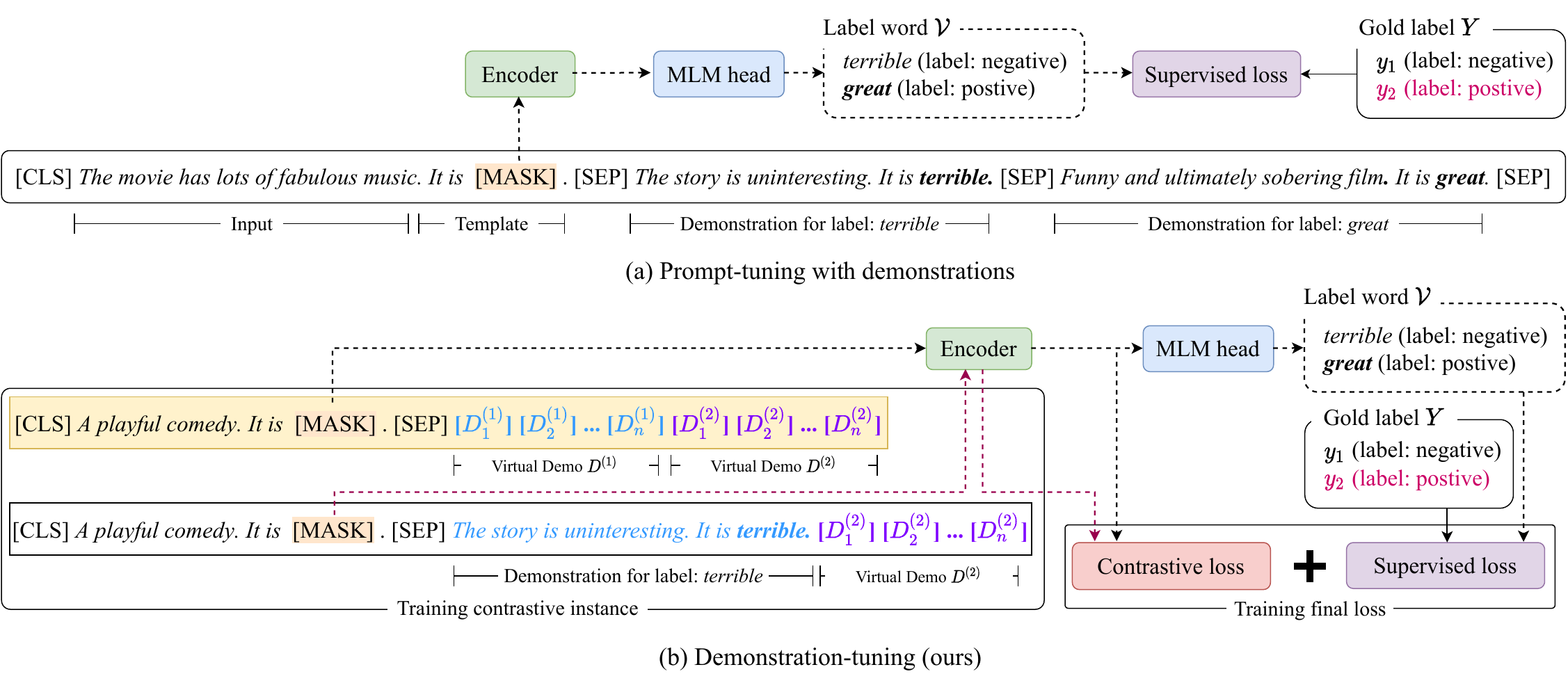}
    \caption{
    An illustration of (a) prompt-tuning with demonstrations, and (b) our proposed contrastive demonstration tuning (demo-tuning).
    Note that we regard the input with virtual demonstration and a random sampled real demonstrations as positive pairs for contrastive learning. 
    }
    \label{fig:overview}
\end{figure*}

\section{Preliminaries}

In this work, we focus on classification tasks in the few-shot setting, including text classification and natural language understanding, where the input $x_{\text{in}}$ is either a sentence $x_{\text{in}}=x_{1}$ or a pair of sentences $x_{\text{in}}=(x_{1},x_{2})$. Here, we let $\mathcal{D}_{\text{train}}=\{(x_{i},y_{i})\}_{i}^{K \times |\mathcal{Y}|}$ denote the training set of a downstream task composed of only $K$ training examples per class, where $\mathcal{Y}$ is label space of the task. 
Given a pre-trained language model comprised of two stages: an encoder $f(\cdot)$ and a classifier $g(\cdot)$ \footnote{In standard fine-tuning, the classifier is a set of randomly initialized parameters $\mathbf{W}_{\text{o}} \in \mathbb{R}^{|\mathcal{Y}| \times d}$ with softmax function.}, we encode the input $x_{\text{in}}$ to a sequence of hidden vectors $\{\mathbf{h}_k \in \mathbb{R}^d\}$ and take the hidden vector $\mathbf{h}_{\texttt{[CLS]}} =f(x_{\text{in}})$ of $\texttt{[CLS]}$ \footnote{For simplicity we will denote the hidden vector $\mathbf{h}_{\texttt{[CLS]}}$ of certain input $x_{i}$ through encoder using $\mathbf{h}_i$.} through classifier to obtain the probability distribution $p\left(y \mid x\right)=g\left(\mathbf{h}_{\texttt{[CLS]}}\right)$ over $y \in \mathcal{Y}$.

\paragraph{Prompt-based Fine-tuning} Prompt-based fine-tuning \cite{DBLP:conf/eacl/SchickS21, DBLP:conf/acl/GaoFC20} is an efficient work by designing cloze-style template $\mathcal{T}$ and verbalizer $\mathcal{M} \colon \mathcal{Y} \rightarrow \mathcal{V} $ mapping task labels to individual words from vocabulary $\mathcal{V}$ of pre-trained language model to fill the gap between masked LM objective of pre-trained language model and downstream fine-tuning objective.
\paragraph{Template} In prompt-based fine-tuning paradigm, template $\mathcal{T}$ is mainly comprised of inputs $x_{\text{in}}$ and a prompt $P=[P_i]^m_{i}$, where the prompt could be a series of discrete tokens \cite{DBLP:conf/eacl/SchickS21} or continual pseudo tokens \cite{DBLP:journals/corr/abs-2103-10385}. For instance, in the sentiment analysis task (see Figure \ref{fig:overview}), a template with handcraft prompt may be: $\mathcal{T}(x) = \texttt{[CLS]} x_1 \textit{, It was} \texttt{[MASK]} \textit{.} \texttt{[SEP]}$ where "\textit{It was ... .}" is prompt and $\texttt{[MASK]}$ is target which cast classification task as a language modeling task.
\paragraph{Verbalizer}A verbalizer $\mathcal{M}$ defines a mapping of label tokens from label space of a specific task. In Figure \ref{fig:overview}a, the verbalizer maps "\textit{negative/postive}" to "\textit{terrible/great}". In this way, we could re-use the output weight $W_{v} \in \mathbb{R}^{d \times |\mathcal{V}|}$ refered \textit{MLM head} used in pre-training and model the probability of predicting token $\mathcal{M}\left(y\right) \in \mathcal{V}$ as $p\left(y \mid x\right) = g\left(\mathbf{h_{\texttt{[MASK]}}}\right)$ on hidden vector $\mathbf{h}_{\texttt{[MASK]}}$.

\paragraph{Demonstration} Let $\mathcal{D}_{\text{train}}^{c}$ be the subset of all examples of class $c$. 
We sample demonstrations $d_{c}=(x_{\text{in}}^{(c)},y^{(c)}) \in \mathcal{D}_{\text{train}}^{c}$  and convert it to $\mathcal{T}(x_{\text{in}}^{(c)}, y^{(c)})$ in which $\texttt{[MASK]}$ is replaced by $\mathcal{M}(y^{(c)})$. 
We then combine the original template $\mathcal{T}$ with templates above in all classes to form $\mathcal{T}^{*}(x_{\text{in}})$, which will be used as a template during prompt-based tuning and inference (See Figure \ref{fig:overview}).

\section{Contrastive Demonstration Tuning}
\label{method}
In this work, we focus on how to learn a compact and differentiable \textbf{virtual demonstration} to serve as prompt augmentation instead of designing specific sampling strategies for demonstration-based learning.
We propose a learning framework based on a contrastive learning approach that can be compatible with the current prompt-based learning paradigm. This section introduces the concepts of \textit{contrastive demonstration tuning}  (Demo-tuning) and provides details of this approach.

\paragraph{Virtual Demonstration} Let $[D_{i}^{(c)}]_{i}^{n}$ refer to the virtual demonstration of the $c^{\text{th}}$ class where $n$ is a hyper-parameter to set the length of virtual demonstration, which is far less than the length of real demonstration. For instance, given a template of binary classification task (see Figure \ref{fig:overview}) as:
\begin{equation}
    \widetilde{\mathcal{T}}(x) = \mathcal{T}(x) \oplus [D^{(1)}] \oplus [D^{(2)}]
\end{equation}
where $\oplus$ denotes concatenation of input sequences. $[D^{(1)}]$ and $[D^{(2)}]$ respectively denote the virtual  demonstrations of two classes. Virtual demonstrations could be so flexible that can be integrated to wide variety of prompt learning approaches \cite{DBLP:journals/corr/abs-2103-10385, DBLP:conf/emnlp/LesterAC21}. 

Next, we study how to obtain the optimal virtual demonstrations, which are initialized as a series of pseudo tokens at the start of fine-tuning. To address this challenging problem, we propose to use contrastive learning, which aims to obtain effective representation by pulling semantically close neighbors together. Intuitively, we believe the optimal virtual demonstrations may be analogous with ``prototype'' \cite{DBLP:conf/nips/SnellSZ17}, the representative for corresponding class, and we will discuss in \S \ref{discussion}.

\paragraph{Positive Instances} A key element of contrastive learning is how to construct reasonable  $\left(x_{\text{in}},x_{\text{in}}^{+}\right)$ pairs. Here, we design a new template $\widetilde{\mathcal{T}}^{+}(x)$ based on template $\widetilde{\mathcal{T}}(x)$ by randomly replacing one of virtual demonstrations $[D^{(c)}]$ with real demonstration $d_c$ as shown in the Figure \ref{fig:overview}b:
\begin{equation}
    \widetilde{\mathcal{T}}^{+}(x) = \mathcal{T}(x) \oplus \mathcal{T}(x^{(1)}_{\text{in}},y^{(1)}) \oplus [D^{(2)}]
\end{equation}
where $[D^{(1)}]$ is replaced with a demonstration $d_1$ of class ``\textit{terrible}''. Using this template, we could convert input $x_{\text{in}}$ to corresponding positive example $x_{\text{in}}^{+}$, i.e., $\left(\widetilde{\mathcal{T}}(x_{\text{in}}),\widetilde{\mathcal{T}}^{+}(x_{\text{in}})\right)$ is a positive training instance.
In this way, aligning virtual demonstration $[D^{(c)}]$ with $d_{c}$, the only difference between $x_{\text{in}}$ and $x^{+}_{\text{in}}$,  and pulling representations $(\mathbf{h}_{\text{in}}, \mathbf{h}_{\text{in}}^{+})$ closer in semantic space could effectively alleviate the problem that the existing of terrible or irrelevant demonstration by previous sampling strategies. 

\begin{table*}[t]
\begin{center}
\centering
\resizebox{1.0\textwidth}{!}{%
\begin{tabular}{lccccccc}
\toprule
& \tf{SST-2} & \tf{SST-5} & \tf{MR} & \tf{CR} & \tf{MPQA} & \tf{Subj} &  \tf{TREC} \\
& (acc) & (acc) & (acc) & (acc) & (acc) & (acc) & (acc)\\
\midrule
``GPT-3'' in-context learning & 84.8 (1.3) &	30.6 (0.9) &	80.5 (1.7) & 87.4 (0.8) & 63.8 (2.1) &	53.6 (1.0) &	26.2 (2.4)  \\
Fine-tuning & 81.4 (3.8) & 43.9 (2.0) & 76.9 (5.9) & 75.8 (3.2) & 72.0 (3.8) & 90.8 (1.8) & 88.8 (2.1) \\
LM-BFF (w/ Demo) & 92.6 (0.5) & \tf{50.6 (1.4)} & 86.6 (2.2) & 90.2 (1.2) & \tf{87.0 (1.1)} & 92.3 (0.8) & 87.5 (3.2) \\
P-tuning (w/ Demo) & 92.7 (1.4) & 47.7 (3.3) & 87.5 (1.3) & 90.6 (1.4) & 84.3 (0.8) & 91.4 (1.7) & 88.1 (2.7) \\
\midrule
Demo-tuning (LM-BFF) & \tf{93.2 (0.4)} & 50.1 (0.4) & \tf{87.9 (0.6)} & \tf{91.5 (0.6)} & 85.9 (1.5) & \tf{92.3 (0.6)} & 90.1 (2.7) \\
Demo-tuning (P-tuning) & 92.7 (0.6) & 48.7 (2.0) & 86.4 (1.1) & 91.4 (0.8) & 86.0 (1.6) & 92.0 (0.6) & \tf{90.7 (4.5)}\\
\midrule
& \tf{MNLI} & \tf{MNLI-mm}  & \tf{SNLI} & \tf{QNLI} &  \tf{RTE} & \tf{MRPC} & \tf{QQP}\\
& (acc) & (acc) & (acc) & (acc) & (acc) & (F1) & (F1) \\
\midrule
``GPT-3'' in-context learning & 52.0 (0.7) &	53.4 (0.6) &	47.1 (0.6) &	53.8 (0.4) &	60.4 (1.4) &	45.7 (6.0) & 36.1 (5.2) \\
Fine-tuning & 45.8 (6.4) & 47.8 (6.8) & 48.4 (4.8) & 60.2 (6.5) & 54.4 (3.9) & 76.6 (2.5) & 60.7 (4.3)\\
LM-BFF (w/ Demo) & 70.7 (1.3) & 72.0 (1.2) & \tf{79.7 (1.5)} & 69.2 (1.9) & 68.7 (2.3) & 77.8 (2.0) & 69.8 (1.8) \\
P-tuning (w/ Demo) & 71.0 (2.2) & 70.8 (1.7) & 78.7 (1.5) & 68.2 (2.1) & \tf{70.8 (3.0)} & 75.0 (13.8) & 66.6 (2.9) \\
\midrule
Demo-tuning (LM-BFF) & 71.0 (2.0) & 72.8 (1.5) & 78.7 (1.9) & \tf{73.1 (1.8)} & 70.0 (3.4) & \textbf{78.4 (2.3)} & \tf{70.2 (1.7)} \\
Demo-tuning (P-tuning) & \tf{71.3 (1.3)} & \tf{73.1 (1.9)} & 76.4 (1.7) & 71.6 (3.0) & 69.8 (4.6) & 78.4 (4.4) & 68.9 (2.9) \\
\bottomrule
\end{tabular}}
\end{center}
\caption{Comparison of performance of our approach with several baselines across 14 text classification tasks in few-shot setting. We report mean (and standard deviation) results of 5 random seeds. LM-BFF (w/ Demo) and P-tuning (w/ Demo): prompt-tuning methods (LM-BFF and P-tuning) using demonstration in context with manual template used in \citet{DBLP:conf/acl/GaoFC20}. Demo-tuning (LM-BFF) and Demo-tuning (P-tuning): Our proposed approach respectively based on LM-BFF and P-tuning.}
\label{tab:main_result}
\end{table*}

\paragraph{Optimization} Similar to \citet{DBLP:conf/icml/ChenK0H20}, we can randomly sample a minibatch of $N$ examples from $\mathcal{D}_{\text{train}}$ to construct positive pairs $\{(x_{i},x_{i}^{+})\}_{i=1}^{N}$  and take a cross-entropy objective with in-batch negatives for $(x_{i},x_{i}^{+})$:
\begin{equation}
    \ell_i = -\log \frac{\exp ({\mathrm{sim}(\mathbf{h}_i, \mathbf{h}^+_i)/\tau})}{\sum_{j=1}^N \exp({\mathrm{sim}(\mathbf{h}_i, \mathbf{h}_j^+)/\tau})}
    \label{eq:cl_neg}
\end{equation}
where $\tau$ denotes a temperature parameter and $\mathrm{sim}(\mathbf{h_i},\mathbf{h_j})$ is the cosine similarity $\frac{\mathbf{h}_i^{\mathrm{T}}\mathbf{h}_j}{\|\mathbf{h}_i\| \cdot \|\mathbf{h}_j \|}$. The negative pairs are composed of two different examples with the same demonstration in a minibatch.

In this work, we also explore a simple contrastive framework without negative pairs\footnote{This is the default contrastive learning method in all experiments.} similar to recent \textit{non-contrastive} self-supervised learning \cite{DBLP:conf/nips/GrillSATRBDPGAP20}.
Regarding the difficulty to find a appropriate negative pair in semantic space for NLP, specially in few-shot setting, we only construct positive pairs and define the following mean squared error between $\mathbf{h}_i$ and $\mathbf{h}_i^{+}$ with $\ell_2\text{-normalization}$,
\begin{equation}
    \ell_i = {\| \mathbf{h}_i - \mathbf{h}^{+}_i \|}_{2}^{2} = 2 - 2 \cdot \frac{\mathbf{h}_i^{\mathrm{T}}\mathbf{h}_i^{+}}{\|\mathbf{h}_i\|_{2} \cdot \|\mathbf{h}_i^{+} \|_{2}}
    \label{eq:contrastive_loss}
\end{equation}
where $\mathbf{h}_i$ and $\mathbf{h}_i^{+}$ are obtained through encoder $f(\cdot)$ in the same state different from \citet{DBLP:conf/nips/GrillSATRBDPGAP20} which encodes $x_i$ and $x_i^{+}$ through two networks in the different states (online network and target network).

When supervised examples $\mathcal{D}_{\text{train}}$ are available, the pre-trained language model could be fine-tuned to minimize the joint objective comprised of cross-entropy and contrastive objective of Eq. (\ref{eq:contrastive_loss}). 
In this way, during inference, we can concatenate the input $x_{in}$ with trained virtual demonstrations in template $\widetilde{\mathcal{T}}(x)$, which does not need to sample real demonstrations. 
Besides, we provide empirical analysis of negative sampling in \S \ref{exp_demo}.

\section{Experiments}

\subsection{Datasets}

To evaluate Demo-tuning, we conduct experiments on 6 tasks from GLUE leaderboard \cite{DBLP:conf/iclr/WangSMHLB19} and 10 other popular classification tasks, including natural language inference (SNLI, MNLI, QNLI, RTE), sentiment classification (SST-2, SST-5, MR, CR, MPQA), paraphrase and similarity (MRPC, QQP) and sentence classification (DBpedia, Subj, TREC, Yahoo! Answers).
The detailed statistics are in Appendix \ref{app:exp}. 

\subsection{Settings}
\paragraph{Evaluation}
During training, we follow the evaluation protocol adopted in \citet{DBLP:conf/acl/GaoFC20} and assume a development set $\mathcal{D}_{\text{dev}}$ for model selection and hyper-parameter tuning, where the size is same with $\mathcal{D}_{\text{train}}$, i.e., $|\mathcal{D}_{\text{dev}}|=|\mathcal{D}_{\text{train}}|$. For every experiment, we measure average performance across 5 different randomly sampled $\mathcal{D}_{\text{train}}$ and $\mathcal{D}_{\text{dev}}$ splits using a fixed set of seeds.

\paragraph{Hyperparameter Selection} 
We implement our framework and reproduce P-tuning by ourselves using PyTorch \cite{DBLP:conf/nips/PaszkeGMLBCKLGA19} and Hugging-Face \cite{DBLP:conf/emnlp/WolfDSCDMCRLFDS20}. The main results of LM-BFF in Table \ref{tab:main_result} are from \citet{DBLP:conf/acl/GaoFC20}. We use $\text{RoBERTa}_{\mathrm{LARGE}}$  \cite{DBLP:journals/corr/abs-1907-11692} as pretrained language model and set $K=16$. 
For the length $n$ of virtual demonstration per class, we select it from candidate set $\{1,2,3,5\}$. 

\subsection{Main Results}
We apply our method to two popular prompt-based tuning techniques, LM-BFF and P-tuning, and compare them to a number of baselines, namely: 
(1) standard fine-tuning in the few-shot setting; 
(2) "GPT-3" in-context learning: zero-shot prediction, which concatenates prompt (e.g., randomly sampled demonstrations); (3) LM-BFF using demonstration in context with a manual template.
(4) P-tuning using demonstration in context with a manual template,  where we do not specifically search the optimal length of continual prompt and fixed the length $m$ to 4 in all tasks.

In Table \ref{tab:main_result}, we report the performance of the baseline approaches and our two variants. 
First, in-context learning could achieve comparable or even higher performance to the standard fine-tuning method and prompt-tuning methods (LM-BFF and P-tuning); using demonstration in context bring consistent improvement in a majority of tasks, which means that demonstration is worth being exploited.

\begin{table}[]
    \centering
    \resizebox{1.0\columnwidth}{!}{
    \begin{tabular}{lcc}
    \toprule
         & \tf{DBpedia} & \tf{Yahoo!} \\
        \midrule
        Fine-tuning & 98.2 (0.1) & 66.4 (1.0) \\
        LM-BFF & 98.1 (0.2) & 66.2 (1.0) \\
        LM-BFF (w/ Demo) & - & - \\
        P-tuning & 98.2 (0.2) & 67.0 (0.8) \\
        \midrule
        Demo-tuning (LM-BFF) & \tf{98.3 (0.1)} & 67.9 (0.8) \\
        Demo-tuning (P-tuning) & \tf{98.3 (0.1)} & \tf{68.4 (1.1)} \\
    \bottomrule
    \end{tabular}}
    \caption{Performance on multi-class sentence classification, DBpedia and Yahoo!. The size of label space $|\mathcal{Y}|$ are respectively 14 and 10. Due to sequence length limitation in pretrained language model, LM-BFF with demonstration-based learning can not be applied here.}
    \label{tab:multi-class}
\end{table}

Second, our approach based on two prompt-based tuning techniques could consistently outperform the vanilla methods. In detail, Demo-tuning based LM-BFF improves the average score by 0.75, compared with LM-BFF with the demonstration in an input context. More importantly, Demo-tuning is flexible and orthogonal to most fine-tuning methods. Here, for evaluating the compatibility, we combine Demo-tuning with P-tuning \cite{DBLP:journals/corr/abs-2103-10385}, which could lead to a 1.0 average score improvement in total. In this work, we do not  specially design template for P-tuning\footnote{We simply construct template $\mathcal{T}(x)$ for P-tuning as $\texttt{[CLS]} x_1 \texttt{[PROMPT]} \texttt{[MASK]} \texttt{[SEP]}$ in single-sentence tasks and $\texttt{[CLS]} x_1 \textit{,} \texttt{[MASK]} \textit{? } x_2 \texttt{[PROMPT]} \texttt{[SEP]}$ in sentence pair tasks, where $\texttt{[PROMPT]}$ denotes continual prompt.}. Although templates for P-tuning and prompt length are suboptimal, we find that Demo-tuning with P-tuning leads to consistent gains in a majority of tasks.

Third, an advantage of our proposed virtual demonstration is that it could be well applied for multi-class sentence classification tasks. Table \ref{tab:multi-class} gives the results of Demo-tuning compared to standard fine-tuning and prompt-based tuning. Due to the limitation of the model's input length, in-context learning and LM-BFF with demonstration could not be applied in this scenario. We notice that while the performance of LM-BFF is worse than fine-tuning, Demo-tuning based on LM-BFF improves the score by 1.7 in Yahoo and achieves a better score compared to fine-tuning.

\begin{table}[]
    \centering
    \resizebox{1.0\columnwidth}{!}{%
    \begin{tabular}{lcccc}
    \toprule
        & \tf{SST-2} & \tf{TREC} & \tf{SNLI} & \tf{MRPC} \\
        \midrule
        LM-BFF & 92.7 & 84.8 & 77.2 & 74.5 \\
        \midrule
        Random & 92.3 & 85.6 & 78.8 & 70.9\\
        Filter-based (RoBERTa) & 92.7 & 83.4 & 79.5 & 76.6 \\
        Filter-based (SBERT) & 92.6 & 87.5 & \tf{79.7} & 77.8 \\
        \midrule
        Virtual Demo (w/ Mean) & 90.9 & 85.9 & 75.3 & 66.4 \\
        Virtual Demo (w/ CL) & \tf{93.2} & \tf{90.7} & 78.7 & \tf{78.4} \\
        \bottomrule
    \end{tabular}}
    \caption{Impact of demonstration sampling strategies. Random: uniform sampling from each class. Filter-based: filtered sampling strategy proposed in \citet{DBLP:conf/acl/GaoFC20} respectively based on RoBERTa and SBERT \cite{DBLP:conf/emnlp/ReimersG19}. Virtual Demo (w/ mean): averaing the representations of instances with the same label as virtual demonstration.}
    \label{tab:demo_sample}
\end{table}

\subsection{Analysis of Virtual Demonstration}
\label{exp_demo}
The selection of demonstration is crucial for demonstration-based learning (e.g., in-context learning and LM-BFF with demonstration). Next, we compare and discuss our proposed virtual demonstration with current approaches.
\paragraph{Demonstration Sampling} Table \ref{tab:demo_sample} provides the impact of demonstration sampling strategies. During inference, our proposed virtual demonstration obtained by contrastive learning during training could be an alternative to real demonstrations, which could be viewed as an implicit sampling strategy. We compare our method with previous sampling strategies based on LM-BFF. 

While the performance of uniform demonstration sampling from each class is better than the vanilla LM-BFF in TREC and SNLI, we notice that on the MRPC task, this method causes severe accuracy loss, which is up to 3.6. We think that random sampling is prone to generate irrelevant information in demonstrations. To address the above issue, \citet{DBLP:conf/acl/GaoFC20} utilize RoBERTa or SBERT \cite{DBLP:conf/emnlp/ReimersG19} to select relevant demonstrations to examples. The filter-based sampling strategy could achieve consistent gains in the majority of tasks, which yields the highest improvement with 3.6 on the TREC task. 
We consider that this KNN-style method, which concatenates examples and demonstrations that are semantically close to examples, could promote language models to decipher meaningful patterns. 

Virtual demonstration, an alternative to the real demonstration during inference, i.e., avoiding complex sampling steps, could achieve gains in most tasks.
Besides our proposed method, We design a simple strategy to construct virtual demonstrations via averaging the representations of instances with the same label. We notice that constructing virtual demonstration with simple averaging of instances causes poor performance in most tasks. However, our method with contrastive learning is more predominant than previous approaches.
The only exception is SNLI, which score only is comparable with random sampling.
We hypothesize that this is caused by some confusion issues, which may exist in filter-based strategy regarding semantically closeness among contrastive demonstrations.

\begin{figure}[t]
    \centering
    \includegraphics[width=1.0\columnwidth]{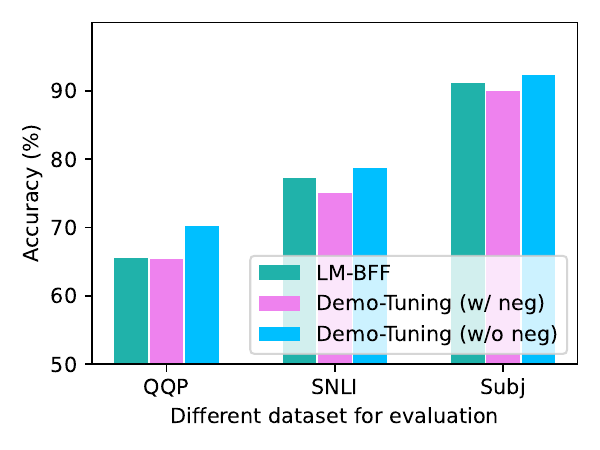}
    \caption{Ablation study on virtual demonstration optimization w/ Vs. w/o negative sampling. Demo-tuning (w/ neg): using conventional contrastive learning with negative samples to optimize virtual demonstration. Demo-tuning (w/o neg): Demo-tuning using our simplified optimization method without negative samples.}
    \label{fig:neg}
\end{figure}

\paragraph{Optimization w/ Vs. w/o Negative Samples} Figure \ref{fig:neg} gives the results of comparison between virtual demonstration optimization with negative sampling and without negative sampling. We conduct experiments with different optimization strategies on 3 tasks. We find that optimizing the objective of Eq.\ref{eq:cl_neg}, i.e., conventional contrastive learning with negative samples, causes dramatically performance degradation, in which the average score is even lower than LM-BFF's. We think there are two possible reasons: (1) In NLP tasks, finding a  semantically reasonable negative pair is difficult, especially in the few-shot setting; (2) Negative pairs may become example-demonstrations pairs without specific limitation, which will cause a certain confusion to model. Moreover, our goal is to obtain optimal virtual demonstrations for downstream tasks. Using contrastive optimization without negative sampling may be a more suitable solution.

\paragraph{Demonstration Length}

\begin{figure}[]
    \centering
    \includegraphics[width=1.0\columnwidth]{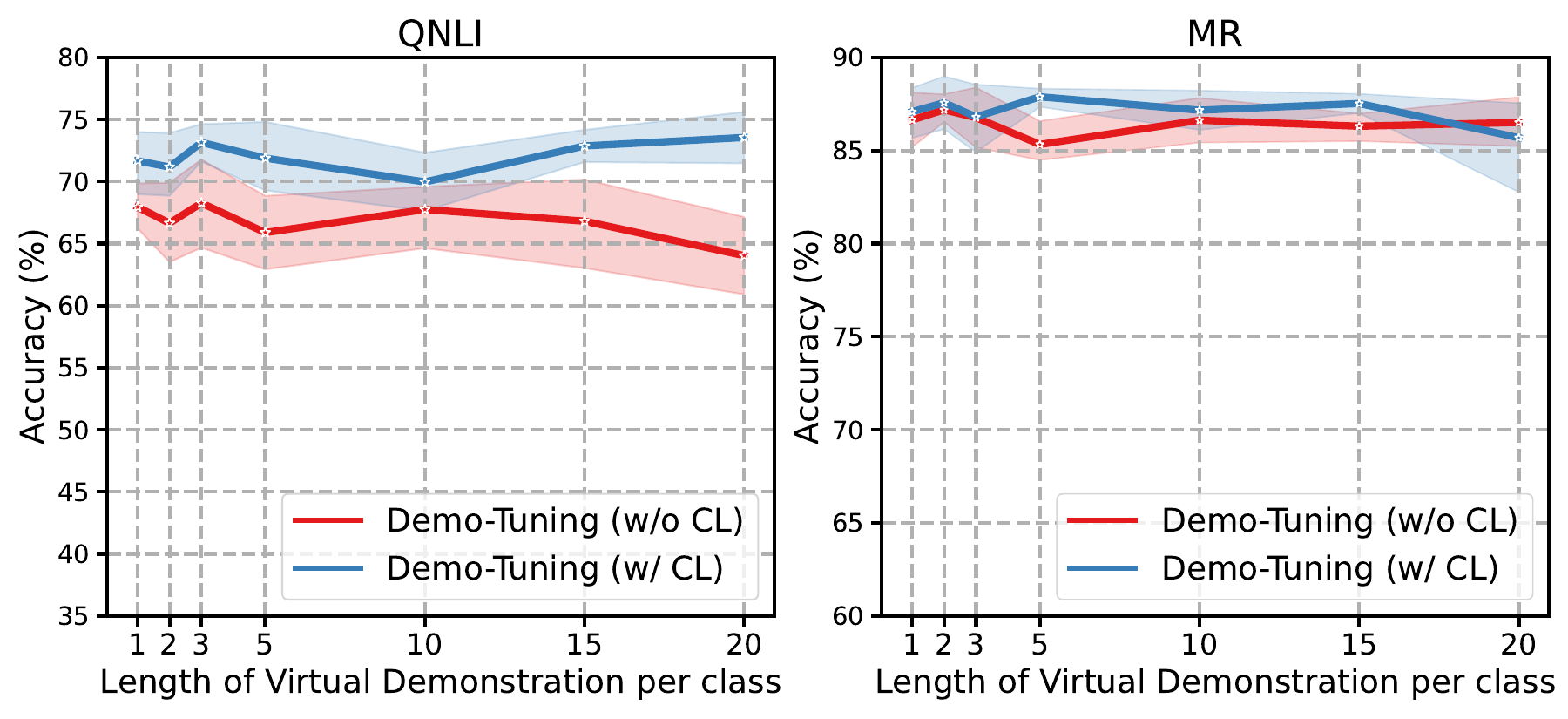}
    \caption{Ablation study on length $n$ of virtual demonstration per class. Demo-tuning (w/o CL): Demo-tuning without contrastive learning (CL), i.e., virtual demonstration will degrade into continual prompt.}
    \label{fig:length}
\end{figure}

Figure \ref{fig:length} shows the ablation study on length $n$ of virtual demonstration per class. We compare Demo-tuning with its variant without contrastive learning in different settings about length $n$. It is noteworthy that without contrastive learning, a virtual demonstration will degrade into a continual prompt. We find that a relatively shorter length (e.g., 2 or 3) could gain stable improvement of performance in QNLI and MR. Oppositely, a larger length (e.g., 20) may decrease the performance. We consider that as the length of virtual demonstration increases, it will introduce more parameters into the model, making it challenging to learn from a small amount of annotated data. Demo-tuning could achieve consistent improvement in different lengths compared to its variant. Hence, we can conclude that \textbf{virtual demonstration optimized by simple contrastive framework plays a different role from continuous prompt}.

\section{Discussion}
\label{discussion}
We will discuss several favorable properties of contrastive demonstration tuning and present some open problems:

\paragraph{Possible Supplement for Parameter-efficient Fine-tuning.}
Previous studies \cite{DBLP:journals/corr/abs-2103-10385,DBLP:conf/acl/LiL20} have demonstrated the effectiveness of prompt-tuning (e.g., P-tuning, Prefix-tuning) as an parameter-efficient fine-tuning methodology for huge PLMs. 
Our approach can serve as a supplement or parameter-efficient fine-tuning via only tuning demonstration with PLM fixed. 
We leave this for future work. 

\paragraph{Relation to Prototype Learning.}
In \S \ref{method}, we note that the optimal virtual demonstrations may be analogous with “prototype” \cite{DBLP:conf/nips/SnellSZ17}, representative for corresponding class.
Our approach may have connections to prototype learning, and further empirical and theoretical analysis should be conducted.

\paragraph{Demonstration as External Knowledge.}
Recall that those concatenated demonstrations are similar to previous studies such as RAG \cite{DBLP:conf/nips/LewisPPPKGKLYR020}, REALM \cite{DBLP:conf/icml/GuuLTPC20} which retrieve and concatenate relevant texts as external knowledge \cite{zhang2022ontoprotein}. 
We think that it is also interesting to investigate novel knowledge injection approaches via demonstration. 

We further discuss a few weaknesses of our method in its current form and look into some
possible avenues for future work. 
On the one hand, our work still suffers from biased/long-tailed label distribution. 
Note that we obtain optimized virtual demonstration via contrastive learning; thus, those virtual demonstrations of classes with many samples may dominate the training stage. 
This limitation might be ameliorated with weighted sampling strategies.
On the other hand, our approach cannot directly handle structure prediction tasks. 
Integrating demonstration with prefix-tuning-based methods may help to mitigate such limitations.

\section{Conclusion and Future Work}

In this work, we propose contrastive demonstration tuning, a simple model-agnostic approach for pre-trained language models, which improves state-of-the-art prompt-tuning performance without the necessity of demonstration selection. 
In the future, we plan to explore the following directions: 
1) studying the connection between virtual demonstration and prototypes and theoretically analyzing the optimal solution of demonstration for prompt-tuning.
2) applying our work to more NLP tasks and trying to adapt to natural language generation.

\section{Limitations}
 
Our contrastive demonstration tuning has limitations. 
Firstly, our model leverages the pre-trained language model; thus, it is necessary to cost  GPU resources. 
Besides, in few-shot settings, the performance gains are still limited with virtual demonstrations learned in only a few training instances.
It is worth studying retrieving relevant context from the internet as ``demonstrations'' to help efficient NLP.

\section*{Acknowledgment}

We want to express gratitude to the anonymous reviewers for their kind comments. 
This work was supported by National Natural Science Foundation of China (No.62206246, 91846204 and U19B2027), Zhejiang Provincial Natural Science Foundation of China (No. LGG22F030011), Ningbo Natural Science Foundation (2021J190), and Yongjiang Talent Introduction Programme (2021A-156-G). Our work was supported by Information Technology Center and State Key Lab of CAD\&CG, ZheJiang University.

\bibliography{custom}
\bibliographystyle{acl_natbib}

\appendix
\section{Datasets}
\label{app:exp}
Table \ref{tab:datasets} provides the dataset evaluated in this work.

\begin{table}[h]
\centering
\resizebox{1.0\columnwidth}{!}{%
\begin{tabular}{lrrrc}
\toprule
\tf{Dataset} & $|\mathcal{Y}|$ & \#Train & \#Test & \tf{Type} \\
\bottomrule
SST-2 & 2 & 6,920 & 872 & sentiment \\
SST-5 & 5 & 8,544 & 2,210 & sentiment \\
MR & 2 & 8,662 & 2,000 & sentiment \\
CR & 2 & 1,775 & 2,000 & sentiment \\
MPQA & 2 & 8,606 & 2,000 & opinion polarity \\
Subj & 2 & 8,000 & 2,000 & subjectivity \\
TREC & 6 & 5,452 & 500 & question cls. \\
DBpedia & 14 & 560,000 & 70,000 & sentence cls. \\
Yahoo! Answers & 10 & 1,400,000 & 60,000 & sentence cls. \\
\midrule
MNLI & 3 & 392,702 & 9,815 & NLI \\
SNLI & 3 &  549,367 & 9,842 & NLI \\
QNLI & 2 & 104,743 & 5,463 & NLI \\
RTE & 2 & 2,490 & 277 & NLI \\
MRPC & 2 & 3,668 & 408 & paraphrase \\
QQP & 2 & 363,846 & 40,431 & paraphrase \\
\bottomrule
\end{tabular}}
\caption{The datasets evaluated in this work. $|\mathcal{Y}|$: the number of classes for classification tasks. Notes that we only sample $\mathcal{D}_{\text{train}}$ and $\mathcal{D}_{\text{dev}}$ of $K \times |\mathcal{Y}|$ examples from the original training data set in our few-shot setting.}
\label{tab:datasets}
\end{table}
\section{Template settings}
\label{app:template}

Table \ref{tab:template} and Table \ref{tab:verbalizer} provides manual templates and verbalizer similar with \citet{DBLP:conf/acl/GaoFC20}. We set the template of demonstration same with example.

\begin{table}
    \centering
    \resizebox{1.0\columnwidth}{!}{%
    \begin{tabular}{ll}
    \toprule
         Template & Tasks \\
    \midrule
         $\texttt{[CLS]} x_1 \textit{, It was} \texttt{[MASK]} \textit{.} \texttt{[SEP]}$  & SST-2, SST-5, MR, CR, MPQA, \\
         & DBpedia, Yahoo! Answers \\
         $\texttt{[CLS]} x_1 \textit{, This is} \texttt{[MASK]} \textit{.} \texttt{[SEP]}$  & Subj \\
         $\texttt{[CLS]} \texttt{[MASK]} \textit{: } x_1 \texttt{[SEP]}$  & TREC \\
    \midrule
        $\texttt{[CLS]} x_1 \textit{?} \texttt{[MASK]} \textit{,} x_2 \texttt{[SEP]}$ & MNLI, SNLI, QNLI, RTE \\
        $\texttt{[CLS]} x_1 \texttt{[MASK]} \textit{,} x_2 \texttt{[SEP]}$ & MRPC, QQP \\
    \bottomrule
    \end{tabular}}
    \caption{Templates for all tasks evaluated in our work.}
    \label{tab:template}
\end{table}

\begin{table}
    \centering
    \resizebox{1.0\columnwidth}{!}{%
    \begin{tabular}{cl}
    \toprule
        Task & Verbalizer \\
    \midrule
       SST-2  &  incorrect/correct \\
       SST-5 & terrible/bad/okay/good/great \\
       MR & terrible/great \\
       CR & terrible/great \\
       MPQA & terrible/great \\
       Subj & subjective/objective \\
       TREC & Description/Entity/Expression/ \\
       & Human/Location/Number \\
       DBpedia & company/institution/artist/athlete/ \\
       & office/holder/transportation/building/ \\
       & place/village/animal/plant/album/film/ \\
       & written/work \\
       Yahoo! & society/science/health/education/ \\
       & internet/sports/business/entertainment/ \\
       & family/politics\\
    \bottomrule
    \end{tabular}}
    \caption{Verbalizer for all tasks evaluated in our work.}
    \label{tab:verbalizer}
\end{table}

\end{document}